\documentclass[11pt,a4paper]{article}
\usepackage{authblk}
\usepackage{booktabs}
\usepackage[hyperref]{emnlp2018}
\usepackage{graphicx} 
\usepackage{graphics}
\graphicspath{{figs/}}
\usepackage{times}
\usepackage{latexsym}
\usepackage{threeparttable}
\usepackage{balance}

\usepackage{soul}
\setstcolor{red}
\setul{}{.2ex}

\usepackage{url}

\aclfinalcopy

\title{Predicting the Usefulness of Amazon Reviews\\Using Off-The-Shelf Argumentation Mining}

\author[$\dagger$]{\textbf{Marco Passon}}
\author[$\ddagger$]{\textbf{Marco Lippi}}
\author[$\dagger$]{\textbf{Giuseppe Serra}}
\author[$\dagger$]{\textbf{Carlo Tasso}}
\affil[$\dagger$]{Universit\`a degli Studi di Udine}
\affil[$\ddagger$]{Universit\`a degli Studi di Modena e Reggio Emilia}
\affil[ ]{\tt marco.passon@spes.uniud.it}
\affil[ ]{\tt marco.lippi@unimore.it}
\affil[ ]{\tt \{giuseppe.serra, carlo.tasso\}@uniud.it} 

\date{}

\begin{document}
\maketitle

\begin{abstract}
Internet users generate content at unprecedented rates. Building intelligent systems capable of discriminating useful content within this ocean of information is thus becoming a urgent need. In this paper, we aim to predict the usefulness of Amazon reviews, and to do this we exploit features coming from an off-the-shelf argumentation mining system. We argue that the usefulness of a review, in fact, is strictly related to its argumentative content, whereas the use of an already trained system avoids the costly need of relabeling a novel dataset. Results obtained on a large publicly available corpus support this hypothesis.
\end{abstract}

\section{Introduction}
\label{sec:intro}
In our digital era, reviews affect our everyday decisions. More and more people resort to digital reviews before buying a good or deciding where to eat or stay.
In fact, helpful reviews allow users to grasp more clearly the features of a product they are about to buy, and thus to understand whether it fits their needs. The same can be said for users who want to book hotels or restaurants.

Companies have started to exploit the importance of reviews. For example, when browsing for a specific product, we are usually presented reviews that have been judged helpful by other users. Moreover, we are often given the possibility to sort reviews according to the number of people who judged them as helpful. That said, a review can also be helpful for companies who want to monitor what people think about their brand. Being able to identify helpful reviews has thus many important applications, both for users and for companies, and in multiple domains.

The automatic identification of helpful reviews is not as easy as it may seem, because the review content has to be semantically analyzed. Therefore, this process is traditionally done by asking users for a judgment. 

To overcome this issue, some approaches have been proposed. One of the earliest studies~\citep{kim2006automatically} aims to rank Amazon reviews by their usefulness by training a regressor with a combination of different features extracted from text and metadata of the reviews, as well as features of the product. Similar approaches employ different sets of features~\cite{ngo2012analyzing}, for example including the reputation of reviewers too~\citep{baek2012helpfulness}.
Another significant work~\citep{mudambi2010research} builds a customer model that describes which features of an Amazon review affect its perceived usefulness, and then it uses such features to build a regression model to predict the usefulness, expressed as the percentage of the number of people who judged a review to be useful.
A hybrid regression model~\cite{ngo2014influence} combines text and additional features describing users (recency, frequency, monetary value) to predict the number of people who judged as useful reviews taken from Amazon and Yelp.
A more complete work considers both regression and classification~\cite{Ghose2011}. It proves different hypotheses, starting with expressing the usefulness of an Amazon review as a function of readability and subjectivity cues, and then converting the usefulness, expressed with a continuous value, into a binary usefulness, that is predicting if a review is useful or not useful.

Another recent work~\cite{Liu2017} presents an approach that explores an similar assumption to ours: helpful reviews are typically \textit{argumentative}. In fact, what we hope to read in a review is something that goes beyond plain opinions or sentiment, being rather a collection of reasons and evidence that motivate and support the overall judgment of the product or service that is reviewed. These characteristics are usually captured by an argumentation analysis, and could be automatically detected by an argumentation mining system~\cite{LippiTOIT2016}. The work in~\cite{Liu2017} considers a set of 110 hotel reviews, it presents a complete and manual labeling of the arguments in such reviews, and it exploits such information as additional features for a machine learning classifier that predicts usefulness.
In this paper, instead, we investigate the possibility to predict the usefulness of Amazon reviews by using features coming from an automatic \textit{argumentation mining system}, thus not directly using human-annotated arguments. A preliminary experimental study conducted on a large publicly dataset (117,000 Amazon reviews) confirms that this could be really doable and a very fruitful research direction. 

\section{Background}
\label{sec:background}
Argumentation is the discipline that studies the way in which humans debate and articulate their opinions and beliefs~\citep{WaltonArgAIBook}.
Argumentation mining~\cite{LippiTOIT2016} is a rapidly expanding area, at the cross-road of many research fields, such as computational linguistics, machine learning, artificial intelligence. The main goal of argumentation mining is to automatically extract arguments and their relations from plain textual documents.

Among the many approaches developed in recent years for argumentation mining, based on advanced machine learning and natural language processing techniques, the vast majority is in fact genre-dependent, or domain-dependent, as they exploit information that is highly specific of the application scenario.
Due to the complexity of these tasks, building general systems capable of processing unstructured documents of any genre, and of automatically reconstructing the relations between the arguments contained in them, still remains an open challenge.

In this work, we consider a simple definition of argument, inspired by the work by Douglas Walton~(\citeyear{WaltonArgAIBook}), that is the so-called claim/premise model. A \textit{claim} can be defined as an assertion regarding a certain topic, and it is typically considered as the conclusion of an argument. A \textit{premise} is a piece of evidence that supports the claim, by bringing a contribution in favor of the thesis that is contained within the claim itself.

\section{Methodology}
\label{sec:methodology}
Our goal is to develop a machine learning system capable of predicting the usefulness of a review, by exploiting information related to its argumentative content. In particular, we consider to enrich the features of a standard text classification algorithm with features coming from an argumentation mining system.
To this aim, we use MARGOT~\cite{Lippi2016}, a publicly available argumentation mining system\footnote{\url{http://margot.disi.unibo.it}} that employs the claim/premise model (to our knowledge, there are no other off-the-shelf systems that perform argumentation mining). Two distinct classifiers, based on Tree Kernels~\cite{Moschitti2006} are trained to detect claims and premises (also called evidence), respectively. When processing a document, MARGOT returns two scores for each sentence, one computed by each kernel machine, that are used to predict the presence of a claim or a premise within that sentence (by default, MARGOT uses a decision threshold equal to zero).

Consider for example the following excerpt of a review, where the proposition in italics is identified by MARGOT as a claim:
\begin{quote}
The only jam band I ever listen to now is Cream, simply because they were geniuses. They were geniuses because the spontaneity, melodicism, and \textit{fearlessness in their improvisation has never been equaled in rock}, and rarely so in jazz.
\end{quote}
Clearly, such a review is very informative, since it comments on very specific aspects of the product, bringing motivations that can greatly help users in taking their decisions. Similarly, the following excerpt of another review brings very convincing arguments in favor of an overall positive judgment of the product. In this case, both sentences are classified by MARGOT as argumentative.
\begin{quote}
\textit{The music indeed seems to transcend so many moods that most pianists have a very hard time balancing this act and there is an immense discography of these concertos of disjoint and loosely-knit performances. Pletnev pushes a straightforward bravura approach with lyrical interludes -- and his performance pays off brilliantly.}
\end{quote}

\begin{figure}[tbp]
\centering
\includegraphics[width=\linewidth]{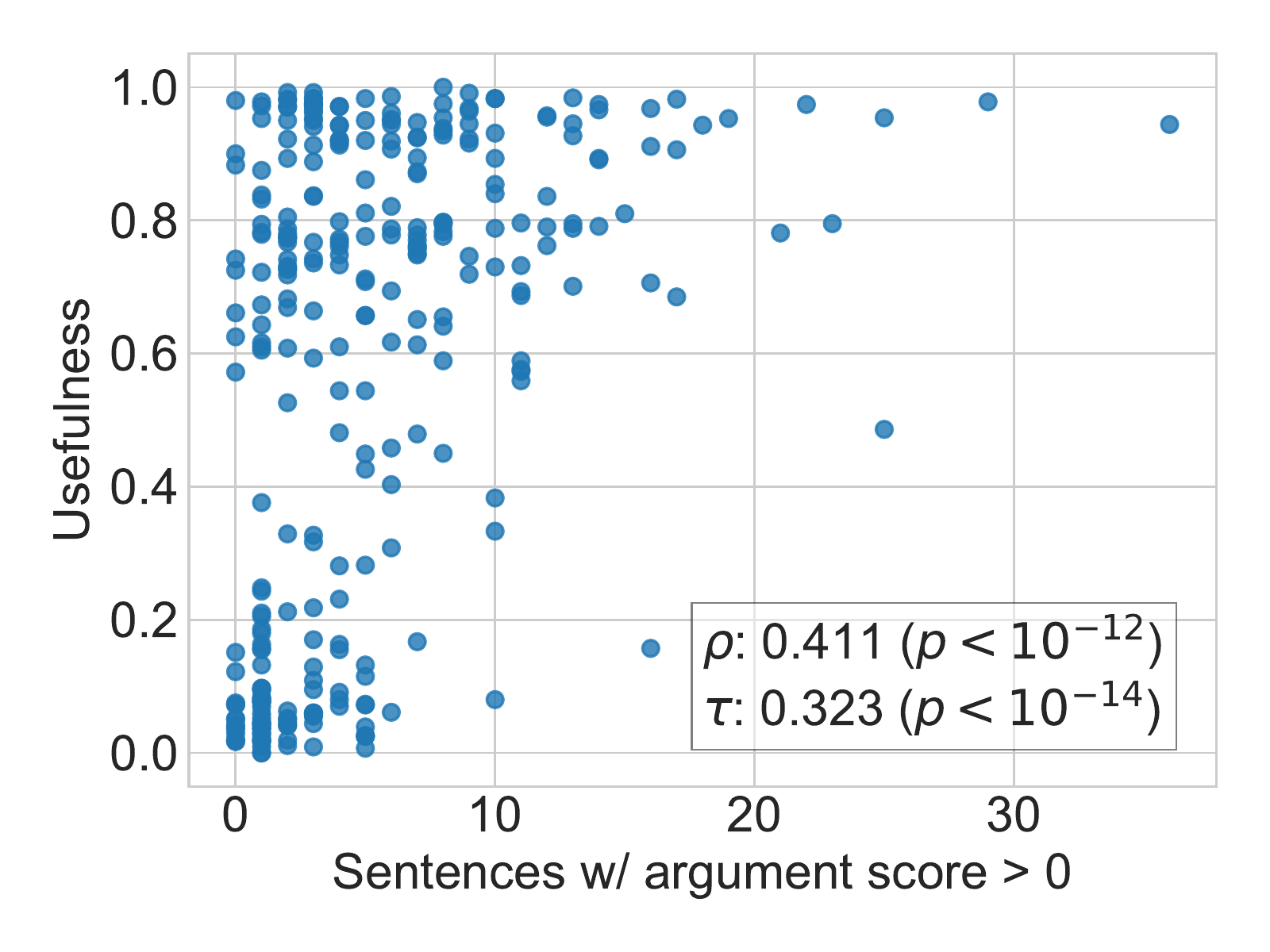}
\caption{Relation between usefulness and number of sentences whose claim or evidence score is above zero for category \lq\lq CDs and Vinyl\rq\rq.\label{fig:CD_threshold}}
\end{figure}

Within this work, we compute simple statistics from the output of MARGOT: the average claim/evidence/argument score, the maximum claim/evidence/argument score, the number and the percentage of sentences whose claim/evidence/argument score is above 0 (that is, the number and the percentage of sentences that contain a claim, an evidence or simply one of those). From a preliminary analysis, in fact, we observed how the presence of arguments within a review is highly informative of its usefulness. Figure~\ref{fig:CD_threshold}, for example, shows the correlation of the number of sentences whose claim or evidence score, according to MARGOT, is above 0, with the usefulness for a subset of 200 reviews in the Amazon category ``CDs and Vinyl''. While it is true that a low number of sentences that contain a claim or an evidence does not necessarily mean that the review is useless, yet the figure shows that a review with a high number of sentences containing a claim or an evidence is most likely a useful review, which confirms our intuition that useful reviews are in fact argumentative. We use these simple statistics as an additional set of features to be used within a standard text classification algorithm, in order to assess whether the presence of argumentative content can help in predicting how useful is a review.

We hereby remark that using MARGOT within this framework is not optimal, because MARGOT was trained on a completely different genre of documents, that is Wikipedia articles. Therefore, we are dealing with a \textit{transfer learning} task, where the argumentation mining system is tested on a different domain with respect to the one it was originality trained on. Using such a classifier adds a challenge to our approach, but it has the advantage of not needing a labeled corpus of argumentative reviews to train a new argumentation mining system from scratch.
Indeed, more sophisticated systems that take into account argumentation could be developed: here, we just want to exploit a straightforward combination of features in order to test our hypothesis.

\section{Experimental Results}
\label{sec:results}
To evaluate the proposed approach we use the public Amazon Reviews dataset~\cite{mcauley2013hidden}, in particular, we worked with the so called \lq\lq 5-core\rq\rq\ subset, that is, a subset of the data in which all users and items have at least five reviews. Each element of this dataset contains a product review and metadata related to it.

Since we aim to predict usefulness, for each review we compute the ratio between the number of people who voted and judged that review as useful, and the total number of people who expressed a judgment about that review. Then, we define useful reviews as the ones whose percentage of usefulness is equal or above 0.7 (that means that at least 70\% of the people who judged a review, judged it as useful), while the remaining are considered not to be useful, and thus they represent our negative class.

The Amazon Review dataset is split into product categories. For our experiments we picked three of them, chosen among those with the highest number of reviews. Our choice has fallen upon the \lq\lq CDs and Vinyl\rq\rq\, \lq\lq Electronics\rq\rq\ and \lq\lq Movies and TV\rq\rq\ categories. We further selected only the reviews having at least 75 rates, in order to assess usefulness on a reasonably large set of samples. Finally, we randomly selected 39,000 reviews for each category, ending up with an almost balanced number of helpful and unhelpful reviews.

Our goal in executing the experiments is to predict whether a review is considered useful, by taking into account either its textual content only, or, additionally, also the argumentation mining data coming from MARGOT. In other words, we are working in a binary classification scenario.

In these experiments we use a stochastic gradient descent classifier\footnote{We used \texttt{SGDClassifier} in \textit{scikit-learn}.} with a hinge loss, which is a classic solution in binary classification tasks. We performed the tuning of the $\alpha$ and $\epsilon$ parameters with a 5-fold cross validation over the training set, and we then used the best model to predict over the test set. From the original set of 39,000 reviews, 50\% of them is used as training set, and the other half as the test set. Each category is treated singularly.

We run experiments both employing a plain Bag-of-Words model, and with TF-IDF features. Both preprocessing variants perform tokenization and stemming\footnote{We used \texttt{Snowball} from python \texttt{nltk} library.} and exclude stopwords and words that do not appear more than five times in the whole training set. To regularize the different magnitude of the features, both textual features and argumentation mining features are normalized using the L2 normalization in all our experiments. Textual and argumentative features are simply concatenated into a single vector.
The performance is measured in terms of accuracy (A), precision (P), recall (R), and $F_1$, as in standard text classification applications.

\begin{table}[tbp]
\centering
\caption{Performance on three Amazon categories using different sets of features: Margot features (M), Bag-of-Words (BoW), Bag-of-Words weighted by TF-IDF (TF-IDF), and combinations thereof.\label{tab:results}}
\begin{threeparttable}
\resizebox{\linewidth}{!}{%
\begin{tabular}{@{}llrrrr@{}}
\toprule
Category              & Data & A & P & R & $F_1$ \\
\midrule
CDs and Vinyl                  & M          & .600    & .544  & .772 & .638   \\
                               & BoW        & .756    & .716  & .769 & .742   \\
                               & BoW + M    & .784    & .744  & \textbf{.799} & .771   \\
                               & TF-IDF     & .769    & .736  & .767 & .752   \\
                               & TF-IDF + M & \textbf{.787}   & \textbf{.751}  & .797 & \textbf{.773}   \\
\midrule
Electronics                    & M          & .583    & .529  & \textbf{.744}  & .618   \\
                               & BoW        & .676    & .639  & .656  & .648   \\
                               & BoW + M    & \textbf{.689}   & .640  & .714  & \textbf{.675}   \\
                               & TF-IDF     & .672    & \textbf{.651}  & .612  & .631   \\
                               & TF-IDF + M & \textbf{.689}    & .649  & .684  & .666   \\
\midrule
Movies and TV                  & M          & .564    & .517  & .792  & .625   \\
                               & BoW        & .745    & .705  & .748  & .726   \\
                               & BoW + M    & .773   & \textbf{.741}  & .767  & .754   \\
                               & TF-IDF     & .757    & .719  & .761  & .740   \\
                               & TF-IDF + M & \textbf{.777}    & .739  & \textbf{.784}  & \textbf{.761}   \\
\bottomrule
\end{tabular}}
\end{threeparttable}
\end{table}

Table~\ref{tab:results} shows that, even using only the features obtained from MARGOT, thus completely ignoring the textual content of the review, the accuracy of the classifier is far above a random baseline.
Moreover, results clearly highlights how the improvement obtained by using argumentative features is consistent across all product categories, both using plain BoW and TF-IDF weighting. 
For the \lq\lq CDs and Vinyl\rq\rq and \lq\lq Electronics\rq\rq categories the difference between the classifier exploiting TF-IDF with MARGOT and the one using TF-IDF only is statistically significant according to a McNemar's test, with $p$-value $< 0.01$. The same holds for the BOW classifier, for the \lq\lq Electronics\rq\rq and \lq\lq Movies and TV\rq\rq categories.

It is interesting to notice that, while the \lq\lq CDs and Vinyl\rq\rq\ and the \lq\lq Movies and TV\rq\rq\ categories have similar performance, even when using textual data only, the category \lq\lq Electronics\rq\rq\ results to be the most difficult to predict. One plausible explanation for this is the heterogeneity of such category, that includes many different types of electronic devices. The other two categories, instead, include more homogeneous products. It would be very interesting to further investigate whether certain product categories result to be more suitable for argumentation studies.

\section{Conclusions}
\label{sec:conclusions}
When reading online reviews of products, restaurants, and hotels, we typically appreciate those that bring motivations and reasons rather than plain opinions. In other words, we often look for \textit{argumentative} reviews.
In this paper, we proposed a first experimental study that aims to show how features coming from an off-the-shelf argumentation mining system can help in predicting whether a given review is useful.

We remark that this is just a preliminary study, which yet opens the doors to several research directions that we aim to investigate in future works. First, we certainly plan to use more advanced machine learning systems, such as deep learning architectures, that have achieved significant results in many applications related to natural language processing. In addition, we aim to address different learning problems, for example moving to multi-class classification, or directly to regression.

The combination of textual and argumentative features exploited in this work was effective in confirming our intuition, but it can certainly be improved. While building a dedicated argumentation mining system for product reviews could require an effort in terms of corpus annotation, we believe that transfer learning here could play a crucial role. Beyond using statistics obtained from the output of an argumentation mining system as an additional input for a second-stage classifier, a unified model combining the two steps could result to be a smart compromise for this kind of application.

\bibliography{emnlp2018}
\balance
\bibliographystyle{acl_natbib_nourl}

\end{document}